\documentclass[letterpaper, 10 pt, conference]{ieeeconf}  

\IEEEoverridecommandlockouts                              

\usepackage{graphicx, tipa}
\usepackage{amsmath,amssymb,amsfonts}
\usepackage[caption=false]{subfig}
\usepackage[ruled,linesnumbered, noend]{algorithm2e}
\usepackage{dirtytalk}
\usepackage{hyperref}
\usepackage{gensymb}
\usepackage{tabularx}
\usepackage{multirow}
\usepackage[short]{optidef}
\usepackage{booktabs}
\usepackage{siunitx}
\usepackage{titlesec}

\newcounter{cases}
\newcounter{subcases}[cases]

\makeatletter
\newcommand{\removelatexerror}{\let\@latex@error\@gobble}
\makeatother
\SetKwComment{Comment}{$\triangleright$\ }{}

\newcommand\Tstrut{\rule{0pt}{2.0ex}}         

\title{\bf \huge 
Quadcopter Trajectory Time Minimization and Robust Collision Avoidance via Optimal Time Allocation
}

\author{Zhefan Xu and Kenji Shimada 
\thanks{Zhefan Xu and Kenji Shimada are with the Department of Mechanical Engineering, Carnegie Mellon University, 5000 Forbes Ave, Pittsburgh, PA, 15213, USA.
        {\tt\small zhefanx@andrew.cmu.edu}}%
}

\begin{document}

\maketitle
\thispagestyle{empty}
\pagestyle{empty}

\begin{abstract}
Autonomous navigation requires robots to generate trajectories for collision avoidance efficiently. Although plenty of previous works have proven successful in generating smooth and spatially collision-free trajectories, their solutions often suffer from suboptimal time efficiency and potential unsafety, particularly when accounting for uncertainties in robot perception and control. To address this issue, this paper presents the Robust Optimal Time Allocation (ROTA) framework. This framework is designed to optimize the time progress of the trajectories temporally, serving as a post-processing tool to enhance trajectory time efficiency and safety under uncertainties. In this study, we begin by formulating a non-convex optimization problem aimed at minimizing trajectory execution time while incorporating constraints on collision probability as the robot approaches obstacles. Subsequently, we introduce the concept of the trajectory braking zone and adopt the chance-constrained formulation for robust collision avoidance in the braking zones. Finally, the non-convex optimization problem is reformulated into a second-order cone programming problem to achieve real-time performance. Through simulations and physical flight experiments, we demonstrate that the proposed approach effectively reduces trajectory execution time while enabling robust collision avoidance in complex environments.
\end{abstract}

\section{Introduction}
With the widespread deployment of unmanned aerial vehicles (UAVs) across diverse indoor applications, extensive research efforts in autonomous navigation have yielded well-established solutions to the challenge of generating collision-free trajectories for quadcopters. Nonetheless, achieving trajectories that are both spatially collision-free and temporally efficient involves the simultaneous consideration of multiple constraints, including obstacle positions, control limits, trajectory smoothness, etc. This complex interplay of constraints can potentially lead to suboptimal time efficiency. In addition, inappropriate trajectory time allocation can let the robot approach the obstacles fast, leading to high motion uncertainty and elevating the risk of collision despite the spatial trajectory being collision-free. As a result, developing a robust and optimal framework for time allocation becomes imperative to enhance both navigation efficiency and safety.  

\begin{figure}
    \vspace{0.25cm}
    \centering
    \includegraphics[scale=0.73]{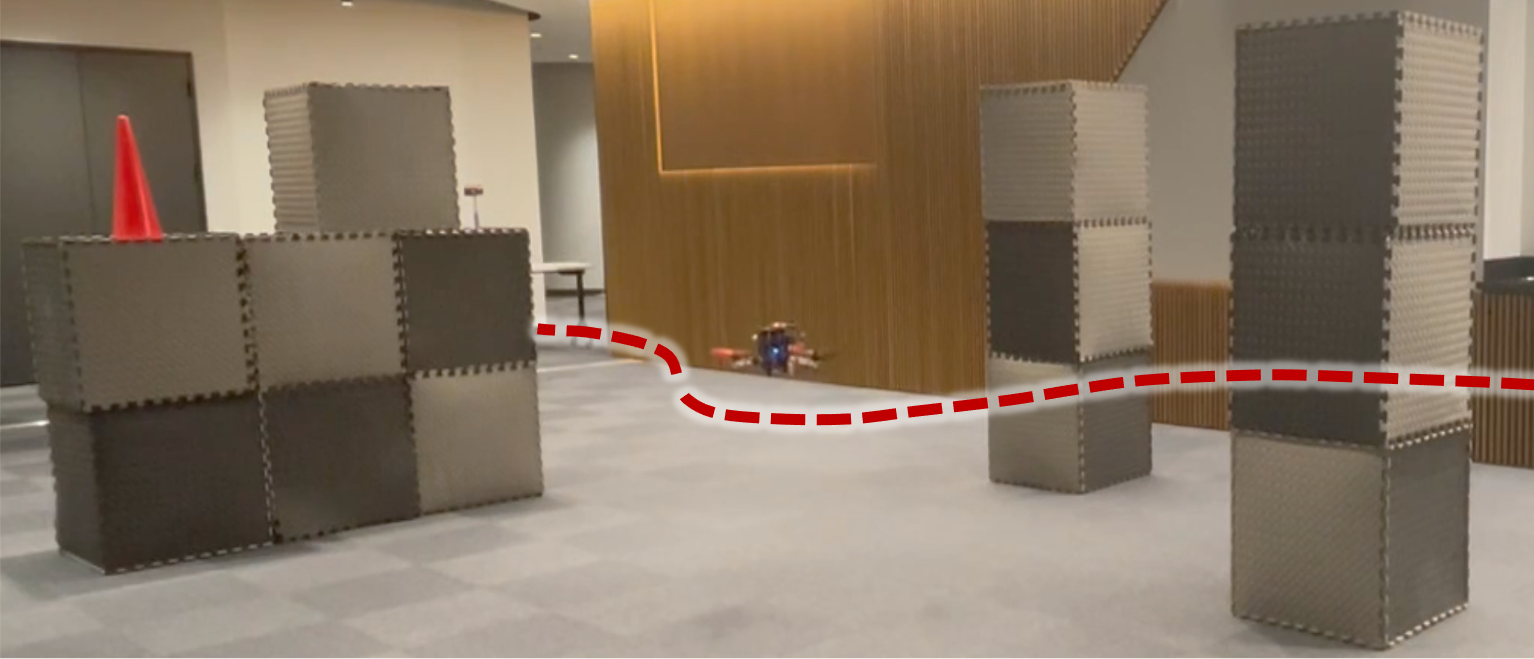}
    \caption{Navigation using the customized quadcopter. The robot follows the trajectory with the optimal time allocation for efficiency and safety.}
    \label{intro figure}
\end{figure}

There exist three primary challenges associated with time allocation within trajectory generation. Firstly, the overall objective is to minimize trajectory execution time. Some trajectory representation and generation techniques, such as MPC-based \cite{ccmpc} and polynomial-based \cite{swarm} methods, rely on pre-defined time allocation, resulting in inflexible timing and non-optimal trajectory execution periods. Secondly, the optimization process must occur in real-time. Although numerical gradient descent methods \cite{minsnap}\cite{bry2015aggressive} have been developed for polynomial-based trajectories, resolving non-convex problems poses constraints on achieving real-time capabilities. Lastly, effective time allocation must ensure trajectory safety marked by uncertainties in perception and control. Even if convex optimization methods \cite{socp}\cite{search} can yield computationally efficient minimum-time solutions, these solutions could potentially induce rapid movements toward obstacles, increasing collision risks amid uncertain conditions.

To solve these issues, this paper proposes the \textbf{R}obust \textbf{O}ptimal \textbf{T}ime \textbf{A}llocation (\textbf{ROTA}) framework. This algorithm serves as a post-processing tool aimed at minimizing execution time while enhancing the safety of quadcopter trajectories under uncertainties. The framework employs a chance-constrained formulation to quantitatively represent a trajectory's collision probabilities and incorporates our proposed trajectory braking zone to establish adaptable velocity constraints for robust collision avoidance. The highly non-convex optimization problem from this framework is subsequently reformulated into a second-order cone programming (SOCP) problem, well-suited for robot onboard computer execution. The novelties and contributions of this work are:

\begin{itemize}
    \item \textbf{Chance-constrained time optimization formulation:} We adopt the chance-constrained scheme to generate time allocation for efficiency and collision avoidance. 
     \item \textbf{Trajectory braking zone determination:} We propose the trajectory braking zone determination algorithm to efficiently find the trajectory deceleration time intervals.
    \item \textbf{Second-order cone programming reformulation:} The original non-convex optimal time allocation problem is reformulated into a second-order cone programming (SOCP) problem to achieve real-time performance. 
\end{itemize}

\section{Related Work}
Time allocation within trajectory generation methods typically falls into two primary categories: algorithmic-based and optimization-based approaches. Given the central theme of enhancing collision avoidance robustness in this study by time allocation, this section will first discuss two categories of time allocation methods and then review related research in path planning and trajectory generation under uncertainty.

The algorithmic-based methods allocate the trajectory time based on a handcrafted algorithm and process. In \cite{algo_kinematics}, the time allocation heuristic based on robot kinematics is applied for each path segment to generate smoother trajectories. Similarly, Jamieson et al. \cite{algo_mapping} design a mapping algorithm to find the proper time allocation for the geometric trajectory to satisfy the kinodynamic limits. Likewise, to limit the maximum control input of the trajectory, Liu et al. \cite{algo_liu} increase the time interval between trajectory positions. In \cite{fast_planner}, an iterative adjustment algorithm is designed to modify the time interval of the non-uniform B-spline curve to generate feasible trajectories. In \cite{swarm}, the time span between each waypoint is increased if the trajectory exceeds the control limits. While algorithmic-based methods exhibit efficiency in generating time allocations, their primary focus often revolves around addressing trajectory feasibility concerns. As a consequence, these methods are unable to guarantee optimality in trajectory time efficiency. There exist several trajectory generation algorithms that use fixed time allocation. The MPC-based methods \cite{ccmpc}\cite{vision_ccmpc}\cite{dpmpc}\cite{tight_cc} require a predefined planning horizon for trajectory generation. Some polynomial-based methods \cite{campos2017hybrid}\cite{burke2020generating}\cite{gao2016online}\cite{bubble_planner} optimize the trajectory based on the fixed time intervals between waypoints.  

The optimization-based methods generate proper time allocations by solving an optimization problem. In quadcopter trajectory planning, Mellinger et al. \cite{minsnap} utilize constrained gradient descent with numerical methods to solve the non-convex time allocation problem to minimize the trajectory snap. Later, Bry et al. \cite{bry2015aggressive} expand upon the minimum-snap concept to simultaneously minimize trajectory snap and execution time. In the work of \cite{roberts2016generating}, the concept of the progressive curve is proposed to minimize the difference with the user-input trajectory progress. Seeking the shortest-time trajectory, Liu et al. \cite{search} apply the search-based planning method and minimize the time for quadcopter motion primitives. To attain spatial and temporal optimality, the alternating minimization method \cite{wang2020alternating} is proposed to solve the spatial-temporal trajectory optimization. This approach's computational efficiency is subsequently improved by transforming the problem into unconstrained optimization \cite{wang2022geometrically}. Within \cite{ego_planner}, the initial spatial trajectory is generated, and feasible time allocation is established via least-square curve fitting. The second-order cone programming for time allocation makes its debut in \cite{first_socp}, later extended to quadcopter trajectory generation in \cite{socp}, demonstrating effectiveness in aggressive flight in complex environments \cite{teach_repeat_replan}.

Recent years have seen plenty of research in quadcopter navigation and collision avoidance for encountering various sources of uncertainties. The introduction of the NanoMap concept in \cite{nanomap} addresses this challenge by storing historical noisy pose estimations, mitigating the adverse impact of pose drift on obstacle avoidance. To achieve robust navigation under sensor uncertainty, Zhang et al. \cite{falcon} model the environment as deterministic and probabilistic known regions. Florence et al. \cite{predefined_manuver} apply a probabilistic model to assess robot maneuvers, enabling fast and robust collision avoidance. The learning-based approach proposed in \cite{learning} tackles camera noise implicitly, facilitating aggressive navigation. In recent research trends, chance-constrained approaches have gained prominence. The linearization technique presented in \cite{ccmpc} significantly enhances computational efficiency and exhibits robust dynamic obstacle avoidance for vision-based quadcopters \cite{vision_ccmpc}. This chance-constrained concept is further expanded to address the simultaneous avoidance of static and dynamic obstacles in subsequent works such as \cite{dpmpc} and \cite{tight_cc}. Benefiting from its efficiency and robustness, our proposed framework also integrates the chance-constrained approach to constrain collision probability for time allocation.

\section{Methodology} 
This section first establishes our time optimization problem (Sec. \ref{problem definition section}) through the introduction of the time mapping function, the concept of the braking zone, and the incorporation of probabilistic collision constraints for time optimization. Following this, we detail the approach for determining the trajectory braking zone (Sec. \ref{braking zone section}). Leveraging the trajectory braking zone, we present linearization techniques to assess the probabilistic collision constraints in Sec. \ref{collision constraint section}. Ultimately, Sec. \ref{socp section} transforms the non-convex optimization problem into a second-order cone programming problem to improve the computational efficiency.

\subsection{Problem Definition \& Optimization Formulation} \label{problem definition section}
\textbf{Time Mapping Function:}  Given a continuous and spatially collision-free trajectory, ${\sigma}_{\text{traj}} \in \mathbb{R}^3$, from an arbitrary trajectory planner, the position at time $\tau$ can be written as:
\begin{equation}
\sigma_{\text{traj}}(\tau) = [\sigma_{\text{traj}}^{\text{x}}(\tau), \sigma_{\text{traj}}^{\text{y}}(\tau), \sigma_{\text{traj}}^{\text{z}}(\tau)]^{T}, \  \tau \in [\tau_{0}, \tau_{f}].
\end{equation}
Note that, for clearer distinction, here we use the symbol $\tau$ to represent the \textbf{trajectory time} variable, and we will also introduce the other time variable $t$ to represent the \textbf{system real time} in the following contexts. To reallocate the time progress of the given continuous trajectory, we need to use a mapping function that maps the system real time to the original trajectory time. We define this mapping function as the time mapping function $h$ with the parameters $\theta$ as: 
\begin{equation}
    \tau = h(t;\theta), \ \ \{h: t \rightarrow \tau \}, \ \ t, \tau \in \mathbb{R},  
\end{equation}
in which the system real time can be mapped to the original trajectory time, and the mapping function is a monotonically increasing function. Thus, the new time reallocated trajectory $\sigma^{r}_{\text{traj}}$ with respect to the real time vairable $t$ is: 
\begin{equation}
    \sigma^{r}_{\text{traj}}(t) = \sigma_{\text{traj}}(\tau) = \sigma_{\text{traj}}(h(t;\theta)),  
\end{equation}
the new corresponding velocity can be written as:
\begin{equation}
    \dot{\sigma}^{r}_{\text{traj}}(t) = \sigma'_{\text{traj}}(\tau) \cdot \dot{h}(t;\theta),  
\end{equation}
and also the reallocated acceleration can be calculated by:
\begin{equation}
    \ddot{\sigma}^{r}_{\text{traj}}(t) = \sigma'_{\text{traj}}(\tau) \cdot \ddot{h}(t;\theta) + \sigma''_{\text{traj}}(\tau) \cdot \dot{h}^{2}(t;\theta),  
\end{equation}
where we use the dot and prime symbol to represent the derivatives to the real time $t$ and the trajectory time $\tau$, respectively. Following the quadcopter controller design in \cite{brescianini2013nonlinear}, we will use the time reallocated position, velocity, and acceleration as control input for trajectory tracking.

\begin{figure}[t] 
    \centering
    \includegraphics[scale=0.82]{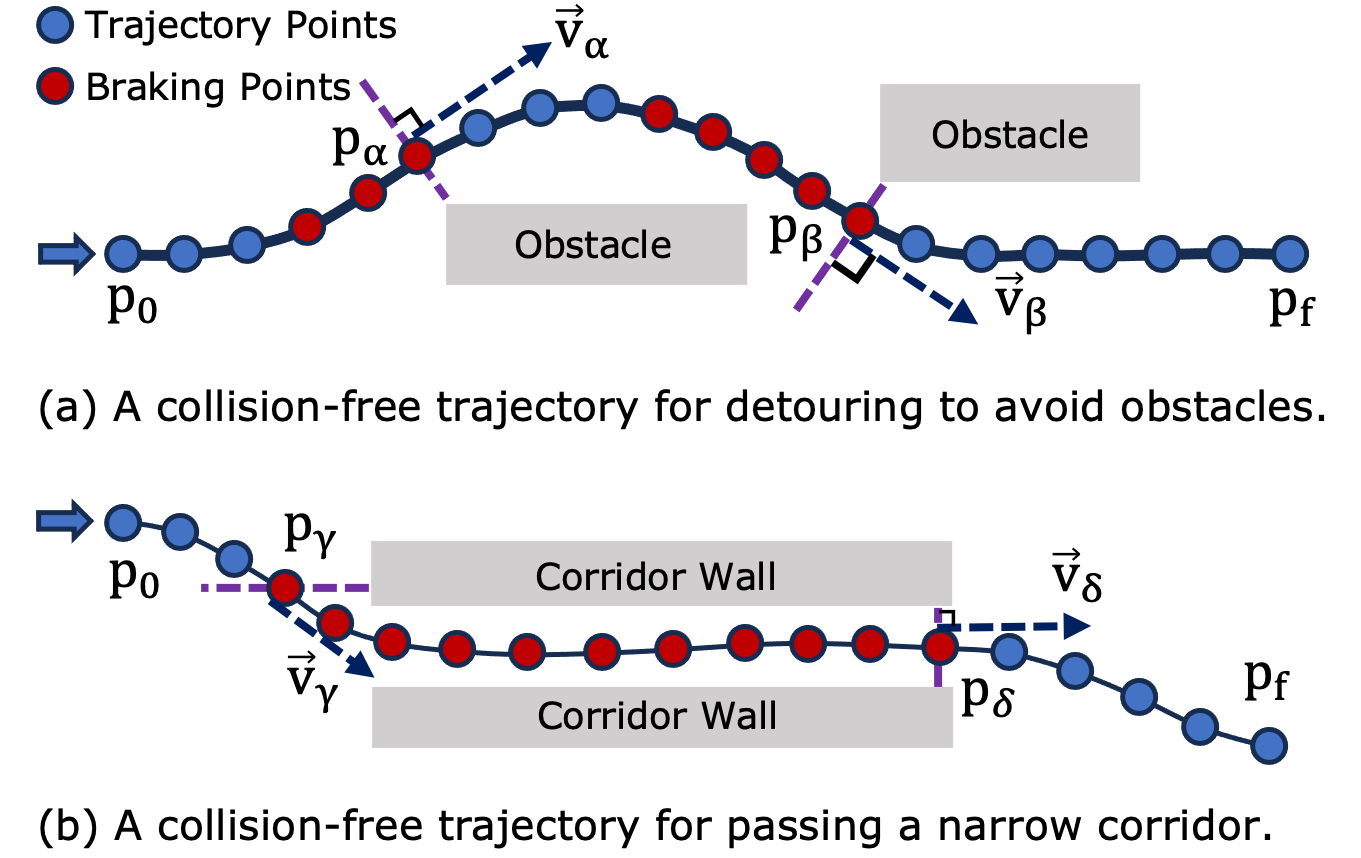}
    \caption{Visualization of the trajectory braking zone. The trajectory's sample positions are depicted as blue and red points, with the red points denoting points within the trajectory braking zone. (a) The robot decelerates to ensure robust collision avoidance when it is approaching obstacles. (b) The braking zone aims at making the robot decelerate to pass a narrow corridor safely.}
    \label{braking zone figure}
\end{figure}

\textbf{Trajectory Braking Zone:} We define the trajectory braking zone as the time intervals that the robot needs potential velocity decrease within the trajectory duration. The intuition behind the trajectory braking zone is to enhance collision avoidance safety and time efficiency by making the robot decelerate when it is approaching obstacles and subsequently re-accelerate upon exiting the braking zone, as shown in Fig. \ref{braking zone figure}. The set of positions $\mathcal{S}_{bz}$ along the trajectory located within the braking zone should adhere to the following conditions: 
\begin{equation} \label{braking points equation}
    \mathcal{S}_{bz} = \{\textbf{p}_{i} \in \mathbb{R}^3 |(\textbf{o}_{i} - \textbf{p}_{i}) \cdot \textbf{v}_{i} \geq 0, ||\textbf{o}_{i} - \textbf{p}_{i}||_{2} \leq d_{\text{th}} \},   
\end{equation}
where $\mathbf{p}_{i}$ and $\mathbf{v}_{i}$ denote the position and velocity of the $i$th sample, respectively. Moreover, $\textbf{o}_{i}$ indicates the nearest obstacle to position $\textbf{p}_{i}$, and $d_{\text{th}}$ represents a user-defined distance threshold. Consequently, the corresponding braking zone $T_{bz}$ with respect to the original trajectory time is:
\begin{equation}
    T_{bz} = [\tau^{bz}_{0}, \tau^{bz}_{f}], \ \sigma_{\text{traj}}(\tau^{bz}_{0}) = \textbf{p}_{0}, \ \sigma_{\text{traj}}(\tau^{bz}_{f}) = \textbf{p}_{f},   
\end{equation}
where $\textbf{p}_{0}$ and $\textbf{p}_{f}$ represent the initial and final positions within the braking position set $\mathcal{S}_{bz}$.
In order to implement braking maneuvers within the braking zones, we dynamically restrict the robot's maximum velocities at each time step, adjusting them according to the proximity of obstacles. This adaptation is achieved through the imposition of collision probability constraints on the trajectory within the braking zones. The determination of the trajectory braking zone and the formulation of detailed constraints will be elaborated upon in Section \ref{braking zone section} and Section \ref{collision constraint section}, respectively.

\textbf{Time Optimization Formulation:} With the introduction of the time mapping function and the trajectory braking zone, we can formulate the time optimization problem. Since we aim at minimizing the total execution time (real time) of the trajectory, the optimization objective can be written in terms of the time mapping function using the chain rule:
\begin{equation}
    \mathcal{J}_{\text{time}}(\theta) = \int^{T_{f}}_{0} 1 dt = \int^{\mathcal{T}}_{0} {\dot{h}(t;\theta)}^{-1}d\tau.
\end{equation}
Following the recommendations from \cite{socp}, we add another objective to improve the time mapping function smoothness:
\begin{equation}
    \mathcal{J}_{\text{smooth}}(\theta) = \int^{\mathcal{T}}_{0} {\ddot{h}(t;\theta)}^{2}d\tau.
\end{equation}
As a result, to achieve the above objectives and ensure robust collision avoidance in the trajectory braking zones, we formulate the chance-constrained time optimization for the parameters $\theta$ of time mapping function $h(t;\theta)$ as:
\begin{mini!}[2]
    {\theta}{\mathcal{J}_{\text{total}}(\theta) = \mathcal{J}_{\text{time}}(\theta) + 
\lambda \cdot \mathcal{J}_{\text{smooth}}(\theta)}{}{} \label{total_objective}
\addConstraint{\text{Pr}(\textbf{p}(t) \in C_{t})\leq \Delta, }{\ \forall \  h(t;\theta) \in [\tau^{bz}_{0}, \tau^{bz}_{f}]}{} \label{chance_constraint}
\addConstraint{\textbf{v}_{\text{min}} \leq \textbf{v}(t) \leq  \textbf{v}_{\text{max}}, \ }{\textbf{a}_{\text{min}} \leq \textbf{a}(t) \leq  \textbf{a}_{\text{max}}}{} \label{control constraint}
\addConstraint{\textbf{p}(t) = {\sigma}^{r}(t),\ \textbf{v}(t) = \dot{\sigma}^{r}(t), \ \textbf{a}(t) = \ddot{\sigma}^{r}(t)}{}{} 
\end{mini!}
\noindent The symbol $C_{t}$ denotes the set of collision positions at time $t$, and $\lambda$ is a parameter for adjusting the weight of objectives. It's worth noting that in Eqn. \ref{chance_constraint}, we require that the trajectory positions within the braking zones must exhibit collision probabilities below a specified threshold denoted as $\Delta$. Considering the implicit representation of the time mapping function $h(t;\theta)$, this optimization task becomes highly non-convex. As a result, Sec. \ref{socp section} will detail our approximate methods for efficiently attaining the solution.

\subsection{Trajectory Braking Zone Determination} \label{braking zone section}
\begin{algorithm}[t] \label{braking zone algorithm}
\caption{Braking Zone Determination} 
\SetAlgoNoLine%
$\mathbb{T} \gets \emptyset$ \Comment*[r]{Initialize an Empty K-D Tree}
$\mathcal{B} \gets \emptyset$ \Comment*[r]{braking points container}
$\mathcal{M} \gets \text{occupancy voxel map}$\;
$\mathcal{\sigma}_{\text{traj}} \gets \text{input collision-free trajectory}$\;
$\mathcal{S}_{\text{local}} \gets \textbf{findLocalRegion}(\mathcal{\sigma}_{\text{traj}})$\; \label{kdtree start}
\For{$\normalfont{\textbf{o}}'_{i}$ \normalfont{\textbf{in}} $\mathcal{S}_{\normalfont{\text{local}}}$}{
    \If{$\mathcal{M}.\normalfont{\textbf{hasCollision}}(\normalfont{\textbf{o}}'_{i})$}{
        $\mathbb{T}.\textbf{insert}(\normalfont{\textbf{o}}'_{i})$ \Comment*[r]{Insert into K-D Tree}
    }
}\label{kdtree end}
\For{$\normalfont{\textbf{p}}_{i}$ \normalfont{\textbf{in}} $\mathcal{\sigma}_{\text{traj}}$}{ \label{start braking points}
    $\textbf{o}_{i} \gets \mathbb{T}.\normalfont{\textbf{nearestNeighbor}}(\normalfont{\textbf{p}}_{i})$\;
    $\textbf{v}_{i} \gets \normalfont{\textbf{getTrajectoryVelocity}}(\textbf{p}_{i})$\;
    \If{$(\normalfont{\textbf{o}}_{i} - \textbf{p}_{i}) \cdot \textbf{v}_{i} \geq 0$ \normalfont{\textbf{and}} $||\normalfont{\textbf{o}}_{i} - \textbf{p}_{i}||_{2} \leq d_{\normalfont{\text{th}}}$}{
        $\mathcal{B}.\normalfont{\textbf{append}}(\textbf{p}_{i})$\;
    }
}\label{end braking points}
$[^{0}T^{\text{raw}}_{bz},...,^{M}T^{\text{raw}}_{bz}] \gets \normalfont{\textbf{convertToTimeIntervals}}(\mathcal{B})$\; \label{merge points}
$[^{0}T_{bz},...,^{N}T_{bz}] \gets \normalfont{\textbf{refine}}([^{0}T^{\text{raw}}_{bz},...,^{M}T^{\text{raw}}_{bz}])$\;
$\textbf{return} \ T_{bz}$\; \label{interval refine}
\end{algorithm}

This section introduces the algorithm (Alg. \ref{braking zone algorithm}) for determining the trajectory braking zones. As discussed in Sec. \ref{problem definition section}, these braking zones correspond to specific time intervals requiring robot deceleration to ensure robust collision avoidance. Moreover, the braking points are defined as the trajectory's sample positions within the braking zones, adhering to the relationships detailed in Eqn. \ref{braking points equation}.

Given an input collision-free trajectory, we first determine the local region, $\mathcal{S}_{\text{local}} \in \mathbb{R}^{3}$, a 3-dimensional bounding box containing the entire trajectory. Utilizing an occupancy voxel map to represent the environment, we iterate through each voxel point $\textbf{o}'_{i}$ and insert the occupied ones into a k-d tree $\mathbb{T}$ (Lines \ref{kdtree start}-\ref{kdtree end}). By constructing this k-d tree, we efficiently store all occupied voxels for efficient nearest obstacle-checking operations. Then, we traverse through each sample position $\textbf{p}_{i}$ on the trajectory to find its nearest obstacle $\textbf{o}_{i}$ (i.e., the occupied voxel). If the trajectory position $\textbf{p}_{i}$ and obstacle $\textbf{o}_{i}$ satisfy the conditions in Eqn. \ref{braking points equation}, the position $\textbf{p}_{i}$ is added into the braking points container $\mathcal{B}$ (Lines \ref{start braking points}-\ref{end braking points}). Subsequently, we can find the corresponding time of braking points in $\mathcal{B}$ and merge them to obtain the raw time intervals (Line \ref{merge points}). The final braking zones are refined from the raw time intervals by merging and removing short intervals (Line \ref{interval refine}).

\subsection{Chance-Constrained Formulation} \label{collision constraint section}
This section details our chance-constrained formulation for the positions within the trajectory braking zone, as described in Eqn. \ref{chance_constraint}. We begin by assuming that the robot positions, $\textbf{p}$, and the obstacle positions $\textbf{o}$ follow the Gaussian distribution as $\textbf{p} \sim \mathcal{N}(\overline{\textbf{p}}, \Sigma_{\textbf{p}})$ and $\textbf{o} \sim \mathcal{N}(\overline{\textbf{o}}, \Sigma_{\textbf{o}})$. So, for $i$th position $\textbf{p}_{i}$ in the braking zone, the collision condition $C_{i}$ with the corresponding nearest obstacle $\textbf{o}_{i}$ can be written as: 
\begin{equation}
C_{i} = \{\textbf{p}_{i} \in \mathbb{R}^3| \begin{Vmatrix} \textbf{p}_{i} - \textbf{o}_{i}  \end{Vmatrix}_{\text{Q}_{c}} \leq 1\}, \label{collision condition}
\end{equation} 
where we assume the safe region of the robot is enclosed by an ellipsoid with semi-axis lengths of $[a, b, c]$ and $\text{Q}_{c}$ is $\normalfont{\textbf{diag}}(\frac{1}{{a}^2}, \frac{1}{{b}^2}, \frac{1}{{c}^2})$. So, under the Gaussian distribution assumption, the collision probability can be calculated by: 
\begin{equation}
\text{Pr}(\textbf{p}_{i} \in C_{i}) = \int_{\begin{Vmatrix} \textbf{p}_{i} - \textbf{o}_{i}  \end{Vmatrix}_{\text{Q}_{c}} \leq 1} p(\textbf{p}_{i} - \textbf{o}_{i})\,d(\textbf{p}_{i} - \textbf{o}_{i}), \label{gaussion_integral}
\end{equation}
which is the integral of the probability density function of the new Gaussian variable $\textbf{p}_{i} - \textbf{o}_{i}$ over the collision region. Since there is no analytical solution to Eqn. \ref{gaussion_integral}, we begin to derive the linearized approximation form by performing the coordinate transform of Eqn. \ref{collision condition} as the follows:
\begin{equation}
C_{i}^{k} = \{\textbf{p}^{r}_{i} \in \mathbb{R}^3 | \begin{Vmatrix} \textbf{p}^{r}_{i} - \textbf{o}_{i}  \end{Vmatrix} \leq 1\}, \label{coordinate transform}
\end{equation} 
where $\textbf{p}^{r}_{i} = \text{Q}_{c}^{\frac{1}{2}} \textbf{p}_{i}$ and the collision condition region is transformed from an ellipsoid to a sphere by eliminating $\text{Q}_{c}$. Subsequently, we can perform the linearization for the transformed collision condition (Eqn. \ref{coordinate transform}) as: 
\begin{equation}
C_{i}^{\text{approx}} = \{\textbf{p}^{r}_{i} \in \mathbb{R}^3 |  \textbf{a}^{T}_{i}(\textbf{p}^{r}_{i} - \textbf{o}_{i}) \leq 1\}, \label{linearized collision condition}
\end{equation}
where $\textbf{a}_{i} = \frac{\overline{\textbf{p}}^{r}_{i} - \overline{\textbf{o}}_{i}}{\begin{Vmatrix} \overline{\textbf{p}}^{r}_{i} - \overline{\textbf{o}}_{i} \end{Vmatrix}}$. With the linearized collision condition, we can approximate the collision probability using:
\begin{equation}
\text{Pr}(\textbf{p}_{i} \in C^{\text{approx}}_{i}) = \text{Pr}(\textbf{a}^{T}_{i}(\textbf{p}^{r}_{i} - \textbf{o}_{i}) \leq 1), 
\end{equation}
where the analytical solution can be further calculated as: 
\begin{equation}
\text{Pr}(\textbf{a}^{T}_{i}(\textbf{p}^{r}_{i} - \textbf{o}_{i}) \leq 1) = \frac{1}{2} + \frac{1}{2}\normalfont{\textbf{erf}}(\frac{1 - \textbf{a}^{T}_{i}(\overline{\textbf{p}}^{r}_{i} - \overline{\textbf{o}}_{i})}{\sqrt{2\textbf{a}^{T} (\Sigma^{r}_{\textbf{p}_{i}}+\Sigma_{\textbf{o}}) \textbf{a}}}), \label{analytical solution}
\end{equation}
and $\Sigma^{r}_{\textbf{p}_{i}} = \text{Q}_{c}^{\frac{1}{2}T} \Sigma_{\textbf{p}_{i}} \text{Q}_{c}^{\frac{1}{2}}$ is the transformed covariance matrix. As a result, if we want to ensure the collision probability is less than the predefined threshold $\Delta$, we have to constrain the covariance matrix $\Sigma_{\textbf{p}_{i}}$. Based on the relation between motion and uncertainty described in \cite{miura2006adaptive}\cite{jin2019obstacle}, a linear function is adopted to describe the relation between the robot velocity $^{i}{v}$ and the standard deviation $^{i}{s}$ along the $i$th axis:
\begin{equation}
    ^{i}{s} = m ^{i}{v} + b, \ \Sigma_{\textbf{p}_{i}} = \normalfont{\textbf{diag}}(^{x}{s}^{2}, ^{y}{s}^{2}, ^{z}{s}^{2}), \label{velocity relation}
\end{equation}
where the coefficients $m$ and $b$ can be found by experiments. Finally, to obtain the velocity limit for each position that satisfies the collision constraint (Eqn. \ref{chance_constraint}) in the trajectory braking zone, we can solve the linear system that combines Eqn. \ref{analytical solution} and \ref{velocity relation} with the probability threshold $\Delta$.

\subsection{Second-Order Cone Programming Formulation} \label{socp section}
To solve the optimization problem (Eqn. \ref{total_objective}) efficiently, we reformulate the optimization into a second-order cone programming (SOCP) problem. Inspired by methods \cite{first_socp}\cite{socp}, we begin by introducing two functions, $\alpha(\tau)$ and $\beta(\tau)$ for representing the time mapping function $h$ with the equations:
\begin{equation}
    \alpha(\tau) = \ddot{h}(t;\theta) = \ddot{\tau}, \ \beta(\tau) = {\dot{h}(t;\theta)}^2 = {\dot{\tau}}^{2}.
\end{equation}
Moreover, since the integration in the objective (Eqn. \ref{total_objective}) does not have an analytical form, we discretize the trajectory by $\delta \tau$ in terms of the original trajectory time $\tau$. The discrete interval number $N = \frac{\tau_{f} - \tau_{0}}{\delta \tau}$ and the functions $\alpha, \beta$ become: 
\begin{equation}
    \alpha = [\alpha_{0}, \alpha_{1},..., \alpha_{N-1}], \ \beta=[\beta_{0}, \beta_{1}, ...,\beta_{N}],
\end{equation}
where we assume $\alpha$ is a piecewise-constant fucntion, and $\beta$ is a piecewise-linear function which should satisfy:
\begin{equation}
    \ddot{h}_{i} = \alpha_{i}, \ \dot{h}_{i} = \frac{\sqrt{\beta_{i}} + \sqrt{\beta_{i+1}}}{2}, \ i \in [0, N].
\end{equation}
Also, $\alpha$ and $\beta$ need to satisfy the following relations:
\begin{equation}
    \alpha_{i} = \frac{\beta_{i+1} - \beta_{i}}{\delta \tau}, \ \beta_{i} > 0.
\end{equation}
So, the objective function is written in terms of $\alpha$ and $\beta$ as:
\begin{equation}
    \mathcal{J}_{\text{total}}(\alpha, \beta) = \sum_{i=0}^{N} (\frac{2}{\sqrt{\beta_{i}}+\sqrt{\beta_{i+1}}} + \lambda \cdot \alpha_{i}) \cdot \delta \tau.
\end{equation}
Since the SOCP requires a linear objective function, two new slack variables $\zeta \in \mathbb{R}^{N+1}$ and $\gamma \in \mathbb{R}^{N}$ are introduced:
\begin{equation}
    \zeta_{i} \leq \sqrt{\beta_{i}} \  \Rightarrow \ 2\cdot(\beta_{i})\cdot(\frac{1}{2}) \geq \zeta^{2}_{i},
\end{equation}
where each $\beta_{i}$ and $\zeta_{i}$ satisfy a rotated cone $\mathbb{Q}_{r}(\beta_{i}, \frac{1}{2}, \zeta_{i})$. So, the new objective function will become the following:
\begin{equation}
    \mathcal{J}_{\text{total}}(\alpha, \zeta) = \sum_{i=0}^{N} (\frac{2}{\zeta_{i}+\zeta_{i+1}} + \lambda \cdot \alpha_{i}) \cdot \delta \tau.
\end{equation}
We further simplify the objective with the variable $\gamma$:
\begin{equation}
    \gamma_{i} \geq \frac{2}{\zeta_{i} + \zeta_{i+1}} \ \Rightarrow \ 2 \cdot (\gamma_{i}) \cdot (\zeta_{i} + \zeta_{i+1}) \geq (2)^{2},
\end{equation}
where variable $\gamma_{i}$ and $\zeta_{i}+\zeta_{i+1}$ forms a rotated cone $\mathbb{Q}_{r}(\gamma_{i}, \zeta_{i}+\zeta_{i+1}, 2)$. So, the final form of the objective function becomes a linear expression in terms of $\alpha$ and $\gamma$:
\begin{equation}
    \mathcal{J}_{\text{total}}(\alpha, \gamma) = \sum_{i=0}^{N} (\gamma_{i} + \lambda \cdot \alpha_{i}) \cdot \delta \tau.
\end{equation}
Next, we can derive the velocity constraints (Eqn. \ref{control constraint}):
\begin{equation}
     ||\sigma'_{\text{traj}}(\tau) \cdot \dot{h}(t;\theta)||_{2}^{2} \leq \textbf{v}^{2}_{\text{max}} \ \Rightarrow \ ||\textbf{v}_{i}||_{2}^{2} \cdot \beta_{i} \leq \textbf{v}^{2}_{\text{max}}.
\end{equation}
With the same linear expression, the collision constraints (Eqn. \ref{chance_constraint}) can also be expressed by the adaptive velocity limits described in Sec. \ref{collision constraint section} at each sample position. Next, the acceleration constraints can be expressed by:
\begin{equation}
    -\textbf{a}_{\textbf{max}} \leq \sigma'_{\text{traj}}(\tau) \cdot \ddot{h}(t;\theta) + \sigma''_{\text{traj}}(\tau) \cdot \dot{h}^{2}(t;\theta) \leq \textbf{a}_{\text{max}} 
\end{equation}
\begin{equation}
    \Rightarrow -\textbf{a}_{\text{max}} \leq \textbf{v}_{i} \cdot \alpha_{i} + \textbf{a}_{i} \cdot \beta_{i} \leq \textbf{a}_{\text{max}}.
\end{equation}
Consequently, we derive both the collision and control constraints into linear forms. Together with the linear objective function and two rotated cones, the optimization becomes a second-order cone programming formulation. The $\alpha$ and $\beta$ solutions can reconstruct the time mapping function $h$.

\section{Result and Discussion}
To evaluate the proposed time allocation performance, we conduct simulation experiments and physical flight tests in different environments. The algorithm is implemented in C++ and ROS with the SOCP solver Mosek. We use the PX4 with Gazebo/ROS for the simulation environments running on Intel i7-12700k@3.8GHz. The physical tests are performed using our customized quadcopter with an Intel RealSense D435i camera and an NVIDIA Orin NX computer. The parameters $m$ and $b$ in Eqn. \ref{velocity relation} are set to $0.2$ and $-0.1$, respectively, based on the experiments, indicating negligible uncertainty under $0.5m/s$ velocity. The threshold collision probability $\Delta$ is set to $0.01 \%$ for each trajectory position. We apply a B-spline-based trajectory planner \cite{xu2023vision} to generate collision-free trajectories and adopt the visual-inertial odometry (VIO) algorithm \cite{vins} for state estimation.

\subsection{Simulation Experiments}
\begin{figure}
    \centering
    \includegraphics[scale=0.50]{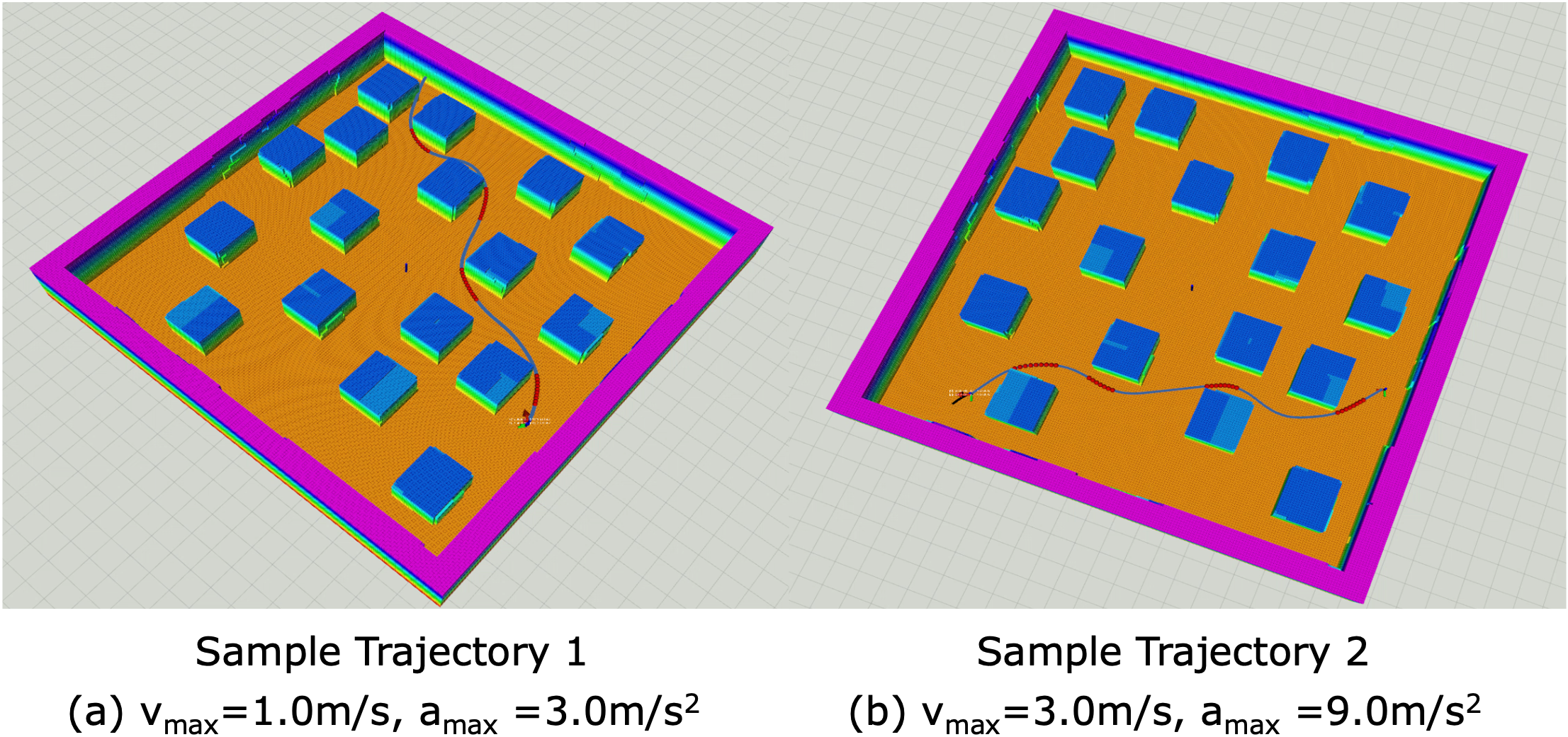}
    \caption{Visualization of two sample trajectories with different control limits. (a) Sample trajectory 1 is planned with a maximum velocity of $1.0m/s$ and acceleration of $3.0m/s^2$. (b) Sample trajectory 2 is generated with a maximum velocity of $3.0m/s$ and acceleration of $9.0m/s^2$.}
    \label{sample image figure}
\end{figure}

\begin{figure}[t] 
    \centering
    \includegraphics[scale=0.49]{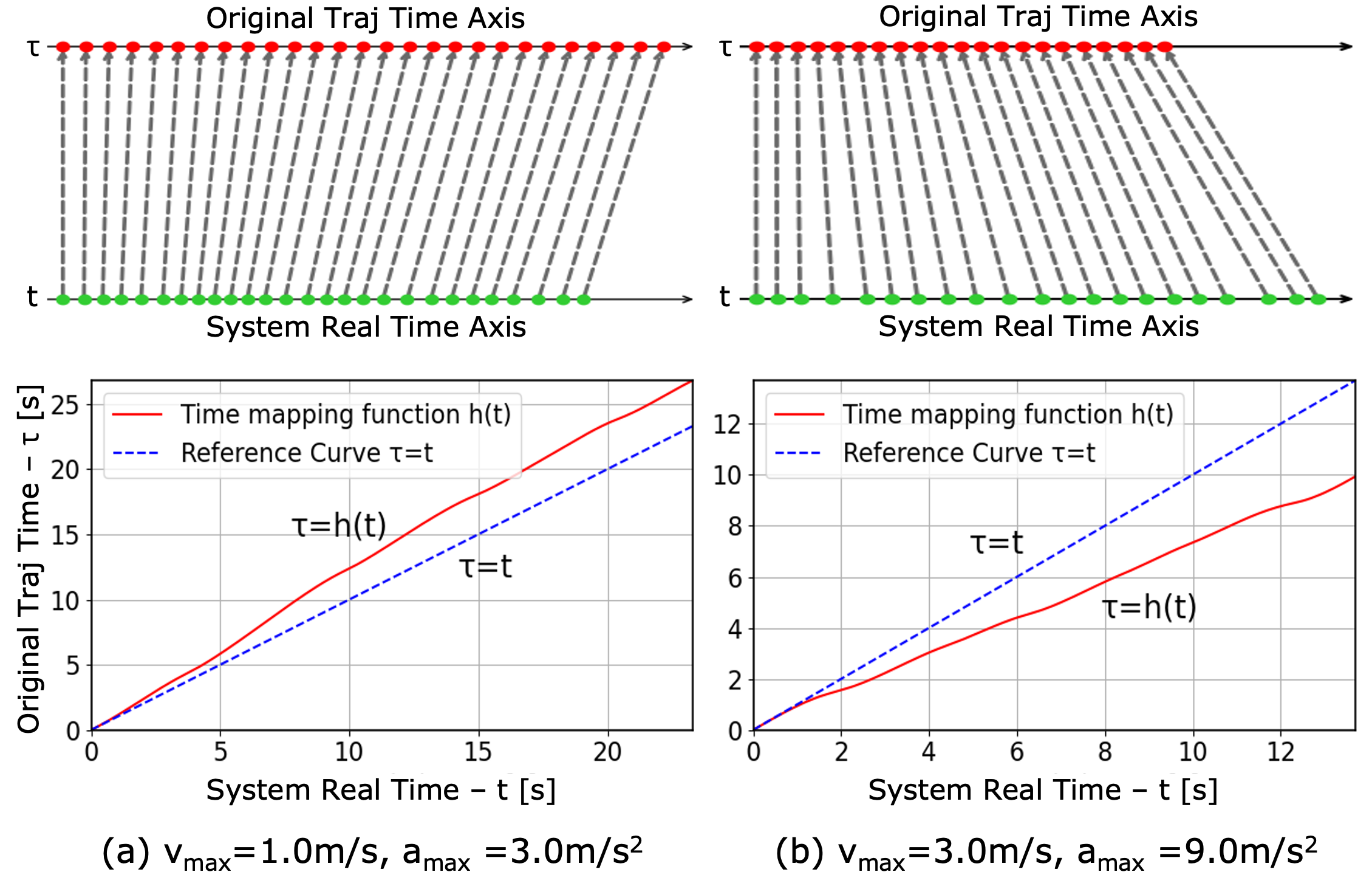}
    \caption{Visualization of the time mapping function from two distinct test samples with different control limits. The top figures depict the mapping relationship between the system real time and the original trajectory time, while the lower figures illustrate the time mapping function curve.}
    \label{time mapping function figure}
\end{figure}

We conduct simulation experiments with different settings to evaluate the performance of the proposed time allocation framework. The sample environments and trajectories with different control limits are shown in Fig. \ref{sample image figure}. Notably, the left trajectory is generated with a maximum velocity of $1.0m/s$ and an acceleration of $3.0m/s^2$ while the right one has a maximum velocity of $3.0m/s$ and a maximum acceleration of $9.0m/s^2$. The resulting time mapping functions of left and right trajectories are visualized in Fig. \ref{time mapping function figure}. The top figures show the mapping relation between the system real and the original trajectory time, and the bottom figures draw the curves of the time mapping functions. From the top left figure, one can see that sample trajectory 1 (with lower control limits) has a shorter trajectory duration after the time allocation, and the time mapping results in a faster-progressing trajectory. This can also be observed in the bottom left curve, as the time mapping function is always higher than the original time curve (the reference curve). However, the algorithm outputs the opposite results for sample trajectory 2 (with higher control limits). The time mapping relation shows a longer execution time and slower trajectory progression, and the time mapping function is lower than the reference curve. Since our objective is to reduce the total time, the time progress of trajectory 1 is optimized for a shorter execution time. However,  because of trajectory 2's higher uncertainties due to its larger velocities, the collision constraints in Eqn. \ref{chance_constraint} slow down the trajectory in the braking zones, increasing the total time.

\begin{figure}[t] 
    \centering
    \includegraphics[scale=0.50]{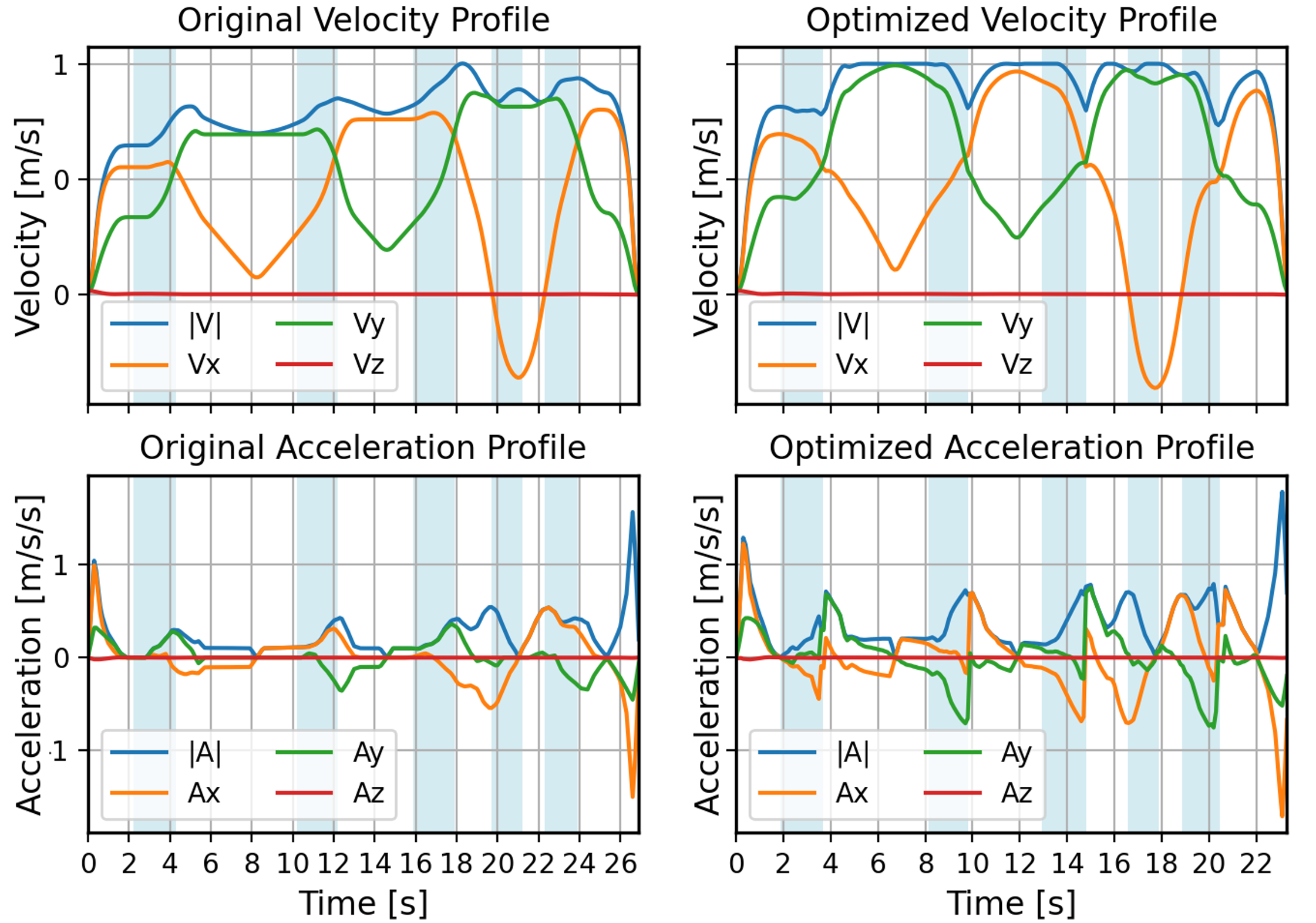}
    \caption{Visualization of the original and optimized velocity and acceleration profiles, where the maximum velocity is set at $1.0m/s$ with an acceleration of $3.0m/s^2$. Notably, the optimized profiles exhibit a reduced trajectory duration. The shaded regions indicate the trajectory braking zones, where the optimized velocity profile shows the robot's deceleration behaviors. }
    \label{trajectory profile figure}
\end{figure}

The comparison of the velocity and acceleration profiles of sample trajectory 1 before and after optimization is visualized in Fig. \ref{trajectory profile figure}. Note that the shaded regions indicate the trajectory braking zones. From the comparison between the origin and optimized profiles, we can see that the optimized profiles reach larger but feasible velocities and accelerations, reducing the total time. Besides, from the braking zone regions, we can see that the optimized velocity profile exhibits the deceleration maneuvers for safe collision avoidance.

To evaluate the effectiveness of safe collision avoidance with the proposed framework, we conduct experiments in environments with different obstacle densities and different control limits. For each setting combination, we run 50 trajectories with and without our time allocation method and record the collision rates. From the results in Table \ref{collision rate result}, we can see that the collision rate will increase with higher obstacle densities and larger velocities without time allocation. However, when the time allocation is applied, the collision rates can be maintained at relatively lower levels. 

\begin{table}[h]
\begin{center}
\caption{Collision rates evaluated from 50 sample trajectories with (left) and without (right) the proposed time allocation.} \label{collision rate result}
\begin{tabular}{ |c | c | c| c |  } 
 \hline

 \multicolumn{4}{|c|}{Collision Rate w/ and w/o the Time Allocation in 20mx20m map} \Tstrut\\
 \hline

 Obstacle Density  & $v=1.0m/s$ & $v=2.0m/s$ & $v=3.0m/s$  \Tstrut\\ 
 \hline

 Low ($0.025/m^2$)  & $0\%$ / $0\%$  & $0\%$ / $0\%$ & $0\%$ / $2\%$ \Tstrut\\ 
 \hline

 Mid ($0.043/m^2$)  & $0\%$ / $2\%$  & $2\%$ / $8\%$ & $4\%$ / $10\%$ \Tstrut\\  
 \hline

 High ($0.063/m^2$) & $0\%$ / $6\%$  & $2\%$ / $12\%$ & $2\%$ / $14\%$ \Tstrut\\ 
 \hline

\end{tabular}
\end{center}
\end{table}

The computation time of the proposed time allocation framework is shown in Table \ref{computation time result}. We record the computation time under different sample time $\delta \tau$ for local ($7m$) and long ($30m$) trajectories. For the local trajectory, the computation time is only around $10$-$30$ms when $\delta \tau \in [0.05, 0.1]$s, bringing negligible latency for local collision avoidance. Moreover, even for the global long trajectory, the computation time can stay below $1$s for a sample time greater than $0.005$s. 

\begin{table}[h]
\begin{center}
\caption{Measurement of Time Allocation Computation Time.} \label{computation time result}
\begin{tabular}{ c  c  c  } 
 \hline

 Sample Time  & Comp. Time ($7m$) & Comp. Time ($30m$) \Tstrut\\ 
 \hline

 $\delta \tau=0.5 \text{s}$  & $0.006\text{s}$  & $0.018\text{s}$ \Tstrut\\ 

 $\delta \tau=0.1 \text{s}$ & $0.017\text{s}$  & $0.049\text{s}$ \Tstrut\\  

 $\delta \tau=0.05 \text{s}$  & $0.026 \text{s}$  & $0.085 \text{s}$ \Tstrut\\ 

 $\delta \tau=0.01 \text{s}$ & $0.093 \text{s}$  & $0.400 \text{s}$ \Tstrut\\  

 $\delta \tau=0.005 \text{s}$  & $0.178 \text{s}$  & $0.840\text{s}$ \Tstrut\\ 

 $\delta \tau=0.001 \text{s}$ & $0.857 \text{s}$  & $5.440 \textbf{s}$ \Tstrut\\  
 \hline
 
\end{tabular}
\end{center}
\end{table}

\subsection{Physical Flight Tests}
\begin{figure}
    \centering
    \includegraphics[scale=0.455]{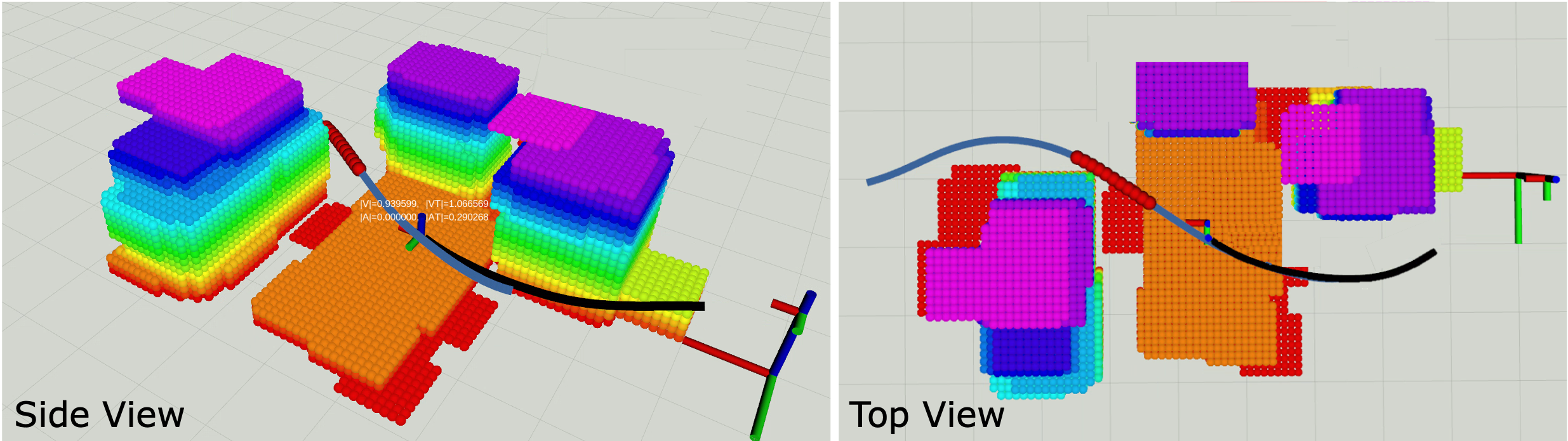}
    \caption{Visualization of the trajectory with the braking zone (red dots) in the physical flight test shown in Fig. \ref{intro figure}. The planned trajectory and the historic trajectory are represented as the blue and black curves. The quadcopter reduces its velocity within the braking zone for safe collision avoidance.}
    \label{experiment rviz figure}
\end{figure}
To verify the performance of the proposed algorithm in the actual robotic system, we conduct the navigation experiments using a customized quadcopter with the velocity limits of $1.0m/s$, $1.5m/s$ and $2.0m/s$ and the acceleration limits of $1.0m/s^2$, $1.5m/s^2$, $2.0m/s^2$ in two different environment settings. The example of one physical flight test (with the limits set to $v_{max}=2.0 m/s$ and $a_{max} = 2.0m/s^2$) is shown in Fig. \ref{intro figure}. The visualization of the trajectory with the braking zone is presented in Fig. \ref{experiment rviz figure}. We use the voxel map mentioned in \cite{mapping} to represent the environment, and the map is inflated by the robot size. In Fig. \ref{experiment rviz figure}, one can see that some parts of the planned trajectory (represented as the blue curve) can be close to obstacles for collision avoidance. Our proposed braking zone determination algorithm can find these risky parts of the trajectory as the trajectory braking zone (shown as the red dots), and our optimal time allocation allows the robot to decrease the velocity for safe collision avoidance.

\section{Conclusion and Future Work}
This paper proposes the Robust Optimal Time Allocation (\textbf{ROTA}) framework to minimize the trajectory time and improve the collision avoidance robustness. Our approach formulates an optimization problem aimed at determining a time mapping function that minimizes time while imposing constraints on collision probability. Notably, we introduce a chance-constrained collision formulation within the trajectory braking zone, enabling the transformation of probability constraints into linear representations. Furthermore, our reformulation of the original non-convex problem into a second-order cone programming (SOCP) problem ensures real-time performance. The simulation results prove the effectiveness of our proposed method in significantly reducing trajectory execution time while concurrently decreasing the likelihood of collisions. Furthermore, physical flight experiments demonstrate the framework's capability to enable safe navigation for quadcopters in complex environments. For future improvement, we will focus on time allocation in the presence of dynamic obstacles with uncertainties.

\bibliographystyle{IEEEtran}
\bibliography{bibliography.bib}

\begin{thebibliography}{10}
\providecommand{\url}[1]{#1}
\csname url@samestyle\endcsname
\providecommand{\newblock}{\relax}
\providecommand{\bibinfo}[2]{#2}
\providecommand{\BIBentrySTDinterwordspacing}{\spaceskip=0pt\relax}
\providecommand{\BIBentryALTinterwordstretchfactor}{4}
\providecommand{\BIBentryALTinterwordspacing}{\spaceskip=\fontdimen2\font plus
\BIBentryALTinterwordstretchfactor\fontdimen3\font minus
  \fontdimen4\font\relax}
\providecommand{\BIBforeignlanguage}[2]{{%
\expandafter\ifx\csname l@#1\endcsname\relax
\typeout{** WARNING: IEEEtran.bst: No hyphenation pattern has been}%
\typeout{** loaded for the language `#1'. Using the pattern for}%
\typeout{** the default language instead.}%
\else
\language=\csname l@#1\endcsname
\fi
#2}}
\providecommand{\BIBdecl}{\relax}
\BIBdecl

\bibitem{ccmpc}
H.~Zhu and J.~Alonso-Mora, ``Chance-constrained collision avoidance for mavs in
  dynamic environments,'' \emph{IEEE Robotics and Automation Letters}, vol.~4,
  no.~2, pp. 776--783, 2019.

\bibitem{swarm}
W.~H{\"o}nig, J.~A. Preiss, T.~S. Kumar, G.~S. Sukhatme, and N.~Ayanian,
  ``Trajectory planning for quadrotor swarms,'' \emph{IEEE Transactions on
  Robotics}, vol.~34, no.~4, pp. 856--869, 2018.

\bibitem{minsnap}
D.~Mellinger and V.~Kumar, ``Minimum snap trajectory generation and control for
  quadrotors,'' in \emph{2011 IEEE international conference on robotics and
  automation}.\hskip 1em plus 0.5em minus 0.4em\relax IEEE, 2011, pp.
  2520--2525.

\bibitem{bry2015aggressive}
A.~Bry, C.~Richter, A.~Bachrach, and N.~Roy, ``Aggressive flight of fixed-wing
  and quadrotor aircraft in dense indoor environments,'' \emph{The
  International Journal of Robotics Research}, vol.~34, no.~7, pp. 969--1002,
  2015.

\bibitem{socp}
F.~Gao, W.~Wu, J.~Pan, B.~Zhou, and S.~Shen, ``Optimal time allocation for
  quadrotor trajectory generation,'' in \emph{2018 IEEE/RSJ International
  Conference on Intelligent Robots and Systems (IROS)}, 2018, pp. 4715--4722.

\bibitem{search}
S.~Liu, N.~Atanasov, K.~Mohta, and V.~Kumar, ``Search-based motion planning for
  quadrotors using linear quadratic minimum time control,'' in \emph{2017
  IEEE/RSJ international conference on intelligent robots and systems
  (IROS)}.\hskip 1em plus 0.5em minus 0.4em\relax IEEE, 2017, pp. 2872--2879.

\bibitem{algo_kinematics}
M.~Shomin and R.~Hollis, ``Fast, dynamic trajectory planning for a dynamically
  stable mobile robot,'' in \emph{2014 IEEE/RSJ International Conference on
  Intelligent Robots and Systems}, 2014, pp. 3636--3641.

\bibitem{algo_mapping}
J.~Jamieson and J.~Biggs, ``Near minimum-time trajectories for quadrotor uavs
  in complex environments,'' in \emph{2016 IEEE/RSJ International Conference on
  Intelligent Robots and Systems (IROS)}.\hskip 1em plus 0.5em minus
  0.4em\relax IEEE, 2016, pp. 1550--1555.

\bibitem{algo_liu}
S.~Liu, M.~Watterson, K.~Mohta, K.~Sun, S.~Bhattacharya, C.~J. Taylor, and
  V.~Kumar, ``Planning dynamically feasible trajectories for quadrotors using
  safe flight corridors in 3-d complex environments,'' \emph{IEEE Robotics and
  Automation Letters}, vol.~2, no.~3, pp. 1688--1695, 2017.

\bibitem{fast_planner}
B.~Zhou, F.~Gao, L.~Wang, C.~Liu, and S.~Shen, ``Robust and efficient quadrotor
  trajectory generation for fast autonomous flight,'' \emph{IEEE Robotics and
  Automation Letters}, vol.~4, no.~4, pp. 3529--3536, 2019.

\bibitem{vision_ccmpc}
J.~Lin, H.~Zhu, and J.~Alonso-Mora, ``Robust vision-based obstacle avoidance
  for micro aerial vehicles in dynamic environments,'' in \emph{2020 IEEE
  International Conference on Robotics and Automation (ICRA)}.\hskip 1em plus
  0.5em minus 0.4em\relax IEEE, 2020, pp. 2682--2688.

\bibitem{dpmpc}
Z.~Xu, D.~Deng, Y.~Dong, and K.~Shimada, ``Dpmpc-planner: A real-time uav
  trajectory planning framework for complex static environments with dynamic
  obstacles,'' in \emph{2022 International Conference on Robotics and
  Automation (ICRA)}, 2022, pp. 250--256.

\bibitem{tight_cc}
T.~Liu, F.~Zhang, F.~Gao, and J.~Pan, ``Tight collision probability for uav
  motion planning in uncertain environment,'' \emph{arXiv preprint
  arXiv:2303.02607}, 2023.

\bibitem{campos2017hybrid}
L.~Campos-Mac{\'\i}as, D.~G{\'o}mez-Guti{\'e}rrez, R.~Aldana-L{\'o}pez,
  R.~de~la Guardia, and J.~I. Parra-Vilchis, ``A hybrid method for online
  trajectory planning of mobile robots in cluttered environments,'' \emph{IEEE
  Robotics and Automation Letters}, vol.~2, no.~2, pp. 935--942, 2017.

\bibitem{burke2020generating}
D.~Burke, A.~Chapman, and I.~Shames, ``Generating minimum-snap quadrotor
  trajectories really fast,'' in \emph{2020 IEEE/RSJ International Conference
  on Intelligent Robots and Systems (IROS)}.\hskip 1em plus 0.5em minus
  0.4em\relax IEEE, 2020, pp. 1487--1492.

\bibitem{gao2016online}
F.~Gao and S.~Shen, ``Online quadrotor trajectory generation and autonomous
  navigation on point clouds,'' in \emph{2016 IEEE International Symposium on
  Safety, Security, and Rescue Robotics (SSRR)}.\hskip 1em plus 0.5em minus
  0.4em\relax IEEE, 2016, pp. 139--146.

\bibitem{bubble_planner}
Y.~Ren, F.~Zhu, W.~Liu, Z.~Wang, Y.~Lin, F.~Gao, and F.~Zhang, ``Bubble
  planner: Planning high-speed smooth quadrotor trajectories using receding
  corridors,'' in \emph{2022 IEEE/RSJ International Conference on Intelligent
  Robots and Systems (IROS)}.\hskip 1em plus 0.5em minus 0.4em\relax IEEE,
  2022, pp. 6332--6339.

\bibitem{roberts2016generating}
M.~Roberts and P.~Hanrahan, ``Generating dynamically feasible trajectories for
  quadrotor cameras,'' \emph{ACM Transactions on Graphics (TOG)}, vol.~35,
  no.~4, pp. 1--11, 2016.

\bibitem{wang2020alternating}
Z.~Wang, X.~Zhou, C.~Xu, J.~Chu, and F.~Gao, ``Alternating minimization based
  trajectory generation for quadrotor aggressive flight,'' \emph{IEEE Robotics
  and Automation Letters}, vol.~5, no.~3, pp. 4836--4843, 2020.

\bibitem{wang2022geometrically}
Z.~Wang, X.~Zhou, C.~Xu, and F.~Gao, ``Geometrically constrained trajectory
  optimization for multicopters,'' \emph{IEEE Transactions on Robotics},
  vol.~38, no.~5, pp. 3259--3278, 2022.

\bibitem{ego_planner}
X.~Zhou, Z.~Wang, H.~Ye, C.~Xu, and F.~Gao, ``Ego-planner: An esdf-free
  gradient-based local planner for quadrotors,'' \emph{IEEE Robotics and
  Automation Letters}, vol.~6, no.~2, pp. 478--485, 2020.

\bibitem{first_socp}
D.~Verscheure, B.~Demeulenaere, J.~Swevers, J.~De~Schutter, and M.~Diehl,
  ``Time-optimal path tracking for robots: A convex optimization approach,''
  \emph{IEEE Transactions on Automatic Control}, vol.~54, no.~10, pp.
  2318--2327, 2009.

\bibitem{teach_repeat_replan}
F.~Gao, L.~Wang, B.~Zhou, X.~Zhou, J.~Pan, and S.~Shen, ``Teach-repeat-replan:
  A complete and robust system for aggressive flight in complex environments,''
  \emph{IEEE Transactions on Robotics}, vol.~36, no.~5, pp. 1526--1545, 2020.

\bibitem{nanomap}
P.~R. Florence, J.~Carter, J.~Ware, and R.~Tedrake, ``Nanomap: Fast,
  uncertainty-aware proximity queries with lazy search over local 3d data,'' in
  \emph{2018 IEEE International Conference on Robotics and Automation
  (ICRA)}.\hskip 1em plus 0.5em minus 0.4em\relax IEEE, 2018, pp. 7631--7638.

\bibitem{falcon}
J.~Zhang, C.~Hu, R.~G. Chadha, and S.~Singh, ``Falco: Fast likelihood-based
  collision avoidance with extension to human-guided navigation,''
  \emph{Journal of Field Robotics}, vol.~37, no.~8, pp. 1300--1313, 2020.

\bibitem{predefined_manuver}
P.~Florence, J.~Carter, and R.~Tedrake, ``Integrated perception and control at
  high speed: Evaluating collision avoidance maneuvers without maps,'' in
  \emph{Algorithmic Foundations of Robotics XII: Proceedings of the Twelfth
  Workshop on the Algorithmic Foundations of Robotics}.\hskip 1em plus 0.5em
  minus 0.4em\relax Springer, 2020, pp. 304--319.

\bibitem{learning}
A.~Loquercio, E.~Kaufmann, R.~Ranftl, M.~M{\"u}ller, V.~Koltun, and
  D.~Scaramuzza, ``Learning high-speed flight in the wild,'' \emph{Science
  Robotics}, vol.~6, no.~59, p. eabg5810, 2021.

\bibitem{brescianini2013nonlinear}
D.~Brescianini, M.~Hehn, and R.~D'Andrea, ``Nonlinear quadrocopter attitude
  control: Technical report,'' ETH Zurich, Tech. Rep., 2013.

\bibitem{miura2006adaptive}
J.~Miura, Y.~Negishi, and Y.~Shirai, ``Adaptive robot speed control by
  considering map and motion uncertainty,'' \emph{Robotics and Autonomous
  Systems}, vol.~54, no.~2, pp. 110--117, 2006.

\bibitem{jin2019obstacle}
J.~Jin and W.~Chung, ``Obstacle avoidance of two-wheel differential robots
  considering the uncertainty of robot motion on the basis of encoder odometry
  information,'' \emph{Sensors}, vol.~19, no.~2, p. 289, 2019.

\bibitem{xu2023vision}
Z.~Xu, Y.~Xiu, X.~Zhan, B.~Chen, and K.~Shimada, ``Vision-aided uav navigation
  and dynamic obstacle avoidance using gradient-based b-spline trajectory
  optimization,'' in \emph{2023 IEEE International Conference on Robotics and
  Automation (ICRA)}.\hskip 1em plus 0.5em minus 0.4em\relax IEEE, 2023, pp.
  1214--1220.

\bibitem{vins}
T.~Qin, P.~Li, and S.~Shen, ``Vins-mono: A robust and versatile monocular
  visual-inertial state estimator,'' \emph{IEEE Transactions on Robotics},
  vol.~34, no.~4, pp. 1004--1020, 2018.

\bibitem{mapping}
Z.~Xu, X.~Zhan, B.~Chen, Y.~Xiu, C.~Yang, and K.~Shimada, ``A real-time dynamic
  obstacle tracking and mapping system for uav navigation and collision
  avoidance with an rgb-d camera,'' in \emph{2023 IEEE International Conference
  on Robotics and Automation (ICRA)}, 2023, pp. 10\,645--10\,651.

\end{thebibliography}

\end{document}